\documentclass{article}

\usepackage[final, nonatbib]{neurips_2020}
\usepackage[utf8]{inputenc} % allow utf-8 input
\usepackage[T1]{fontenc}    % use 8-bit T1 fonts
\usepackage{hyperref}       % hyperlinks
\usepackage{url}            % simple URL typesetting
\usepackage{booktabs}       % professional-quality tables
\usepackage{amsfonts}       % blackboard math symbols
\usepackage{nicefrac}       % compact symbols for 1/2, etc.
\usepackage{microtype}      % microtypography
\usepackage{multirow}
\usepackage{newtxtext,newtxmath}
\usepackage{floatrow}
\usepackage{wrapfig,lipsum,booktabs}
\usepackage[colorinlistoftodos]{todonotes}
\newfloatcommand{capbtabbox}{table}[][\FBwidth]

\usepackage{graphicx}
\usepackage{wrapfig}
\usepackage{subfig}

\title{FAT: Federated Adversarial Training}

\author{%
  \begin{tabular}{c@{\hspace*{.8cm}}c@{\hspace*{.8cm}}c@{\hspace*{.8cm}}c}
  
    {\bf Giulio Zizzo$^\dagger$\thanks{work was done while at IBM Research}}
  & {\bf Ambrish Rawat$^*$} 
  & {\bf Mathieu Sinn$^*$} 
  & {\bf Beat Buesser$^*$}\\[.1cm]
  
  \multicolumn{4}{c}{\normalfont $^\dagger$Department of Computing, Imperial College London}\\[0.1cm]
  \multicolumn{4}{c}{\normalfont $^*$IBM Research}\\[0.1cm]
  \multicolumn{4}{c}{\normalfont \texttt{\{ambrish.rawat,mathsinn,beat.buesser\}@ie.ibm.com}}\\[0.1cm]
  \multicolumn{4}{c}{\normalfont \texttt{g.zizzo17@imperial.ac.uk}}\\[-.3cm]
  \end{tabular}
}

\begin{document}

\maketitle

\begin{abstract}
Federated learning (FL) is one of the most important paradigms addressing privacy and data governance issues in machine learning (ML). Adversarial training has emerged, so far, as the most promising approach against evasion threats on ML models.
In this paper, we take the first known steps towards federated adversarial training (FAT) combining both methods to reduce the threat of evasion during inference while preserving the data privacy during training.
We investigate the effectiveness of the FAT protocol for idealised federated settings using MNIST, Fashion-MNIST, and CIFAR10, and provide first insights on stabilising the training on the LEAF benchmark dataset which specifically emulates a federated learning environment.
We identify challenges with this natural extension of adversarial training with regards to achieved adversarial robustness and further examine the idealised settings in the presence of clients undermining model convergence.
We find that Trimmed Mean and Bulyan defences can be compromised and we were able to subvert Krum with a novel distillation based attack which presents an apparently ``robust'' model to the defender while in fact the model fails to provide robustness against simple attack modifications.
\end{abstract}

\section{Introduction}

Federated learning (FL) is a machine learning paradigm where a central server collects model updates computed locally by participants on their private data and aggregates these updates to train a globally learned model~\cite{mcmahan2017communication}. 
This brings multiple advantages: the clients do not have to share their own data while benefiting from a model trained on all clients' data and sharing the computational effort across all participating devices~\cite{kairouz2019advances}. 
However, models trained this way are as vulnerable to adversarial examples as centrally trained models.

Adversarial examples are inputs to machine learning models that have been imperceptibly manipulated but result in mis-classification with high confidence~\cite{biggio2013evasion, szegedy2013intriguing}.
There is significant research into defences against adversarial examples \cite{papernot2018sok} and adversarial training~\cite{goodfellow2015explaining} has so far achieved the best empirical robustness ~\cite{croce2020reliable}.
Popular protocols of adversarial training~\cite{madry2017towards} train models on mixtures of benign and adversarially perturbed data.
These protocols are sensitive optimisation parameters for generating the adversarial perturbation~\cite{tram2018ensemble, madry2017towards, wong2020fast} and the resulting robustness can vary significantly, especially for large scale datasets~\cite{gupta2020improving,kurakin2017adversarial,zhang2019theoretically,duesterwald2019exploring}.  

Adopting adversarial training to federated learning poses a range of open challenges including poor convergence rates, adoption to non-IID settings, and its interaction with secure aggregation schemes.
This work examines these unexplored challenges and makes headway on the investigation of the interplay of adversarial training with FL.
To the authors' knowledge this is the first study of this kind.
The contributions of this paper are: 
1) evaluating the models trained in a FL setup against inference time evasion attacks, 
2) results and lessons learnt from scaling adversarial training in FL, 
and 3) analysis of adversarial training in the presence of clients that seek to undermine model convergence~\cite{lamport1982byzantine,baruch2019alittle} and its interplay with robust aggregation schemes~\cite{blanchard2017machine,mhamdi2018hidden}. 
As part of this final analysis we devise a novel attacker which introduces gradient masking into the global model. This gives a false sense of security regarding its robustness resulting in models that can be evaded by several different methods.

With potentially millions of users participating in FL~\cite{bonawitz2019towards}, the system must ensure robustness to malicious participants.
However, robust aggregation schemes that apply heavy trimming to the supplied updates may negatively affect the delicate aim of adversarial robustness.
It is thus an open question if both byzantine and adversarial robustness can coexist within a FL system.

\section{Federated Adversarial Training}

The min-max formulation of adversarial training as proposed in~\cite{madry2017towards} consists of alternating between computing adversarial examples using Projected Gradient Descent (PGD) and training the model with the augmented set of these examples.
For learning a model specified with parameters $\theta$ and loss function $\mathcal{L}$, clients that adopt a Federated Adversarial Training (FAT) protocol compute the following gradient for every minibatch of $N$ samples - $\left\{\left(\boldsymbol{x}^{(i)}, y^{(i)}\right)\right\}_{i=1}^{N}$, 
\begin{equation}
    \sum_{i=1}^{K} \nabla_{\theta}\mathcal{L}\left(\boldsymbol{x}_{\mathrm{adv}}^{(i )}, y^{(i)} ; \theta\right)+\sum_{i=K+1}^{N} \nabla_{\theta}\mathcal{L}\left(\boldsymbol{x}^{(i)}, y^{(i)} ; \theta\right).
\end{equation}
$K$ determines the proportion of adversarial samples that are used for the update.
Once the local gradients or weights are computed they are sent to the central server to be aggregated to form a global model.
In the standard setup, the updates are simply averaged~\cite{mcmahan2017communication} and applied to the global model.

As the first step we define an idealised FL setup for standard datasets of MNIST~\cite{mnist}, Fashion-MNIST~\cite{mnist} and CIFAR10~\cite{cifar}, where 51 clients with their equally-sized private partitions of the train and test datasets collaboratively train a model by agreeing to a common training protocol that specifies a choice of optimiser hyperparameters and $K/N$ is set to 0.5.
Furthermore, all clients participate in each communication round of this FL setup.
Following the convention of \cite{madry2017towards}, we report the average accuracy and average robustness accuracy obtained for an untargeted PGD attack.
We see from Table \ref{tab:scale_results} that there is a drop in both regular and adversarial accuracy, however this drop is modest with the worst performance drop corresponding to Fashion-MNIST's adversarial accuracy falling from 80.35\% to 69.4\%.
Having established that in idealised situation adversarial training can be conducted efficiently, we now turn to more realistic datasets with FE-MNIST.

\begin{table}[h!]
    \centering
    \begin{tabular}{ c c c c c c c c}
        \toprule
         Dataset & $L_\infty$ & \multicolumn{2}{c}{PGD Steps} & \multicolumn{2}{c}{Centralised} & \multicolumn{2}{c}{Federated} \\
                 &            &  Num.  &   Size   & Adv. Acc & Acc & Adv. Acc & Acc \\
         \midrule
        MNIST         & 0.3 & 40 & 0.01 & 92.76 & 99.15 & 90.92 & 99.09 \\
        Fashion-MNIST & 0.15 & 15 & 0.01 & 80.35 & 89.33 & 69.41 & 88.67 \\
        CIFAR10       & 8/255 & 10 & 2/255 & 38.58 & 75.20 & 35.59 & 70.29  \\
         \bottomrule
    \end{tabular}
    \caption{Results showing the performance of our models in a traditional centralised training setup compared to a distributed FL scenario. For datasets MNIST, Fashion-MNIST, and CIFAR10 we use 51 clients, mirroring the same configuration that we use in evaluating Byzantine defences}
    \label{tab:scale_results}
\end{table}

\subsection{Practical FAT}

To test the usefulness of FAT in realistic scenarios, we evaluate the protocol for the setup specified in \cite{caldas2018leaf} for the Federated Extended MNIST (FE-MNIST) dataset.
This setup simulates a non-IID scenario comprising of 3500 users where each user owns a set of 28x28 monochromatic images belonging to 62 different classes.
We used the setup specified in~\cite{caldas2018leaf} and train the models for 2500 communication rounds where clients use SGD optimiser with a learning rate of 0.004.
Motivated from the MNIST settings in \cite{madry2017towards} we use an $L_\infty$ budget of 0.3 with a step size of 0.01 for a maximum of 40 steps as PGD parameters and train the model.
When 3 clients are selected during each communication round to perform 1 epoch of adversarial training, we note that the protocol doesn't work out of the box.
The training is unstable and fails to converge for the popular $K/N$ choices of 0.5 and 1.0.

Adopting policies that initially train with smaller proportions of adversarial samples help in guiding the training.
Inspired by similar observations in \cite{gupta2020improving,sinn2019evolutionary,duesterwald2019exploring}, we train for a $K/N$ of 0.1 for initial rounds before jumping to larger proportions.
Two questions are: when should an algorithm make this jump and what value of $K/N$ should it jump to?
We find that a model trained with 0.1 for the first 200 communication rounds followed by training with $K/N$ of 0.8 for the next 2300 rounds exhibits a robust accuracy of 33.69\%.
We analyse the effect of these hyperparameters by comparing the the average accuracy across all clients weighted by their number of test samples (Table \ref{tab:femnist_results}).
We find that prolonging larger proportions beyond 200 rounds significantly slows down the convergence (jump 400 in Figure 1).
Additionally, we compare the performance across protocols that switch to different values of $K/N$ after 200 communication rounds.
We tried $K/N$ of 0.5 - 0.9 and found 0.8 to be optimal with results ranging from 27.68\% to 33.69\%.
Finally, we also note a sensitivity of the optimisation to minibatch size. 
While we report the results for setups where clients adopt a minibatch size of 10 as specified in the original work of \cite{caldas2018leaf}, larger minibatches exhibit slower convergence rates. 

\begin{wraptable}{r}{4.7cm}
\caption{Results on FE-MNIST.}\label{tab:femnist_results}
\begin{tabular}{ c c c}
        \toprule
         Method &  Adv. Acc. & Acc.\\
         \midrule
        FedAvg & 0.0 & 82.61 \\
        FAT-0.5 & 27.68 & 74.81\\
        FAT-0.8 & 33.69 & 72.28\\
        \midrule
        Cent-AT & 37.21 & 82.11 \\
         \bottomrule
    \end{tabular}
\end{wraptable} 

To put the results in perspective, we conduct adversarial training for centralised FE-MNIST with Adam as the optimiser (Cent-AT in Table~\ref{tab:femnist_results}).
We note the same difficulties of convergence with $K/N$ of 0.5 and 1.0 and assist the optimisation by training with the ratio of 0.1 for initial epochs. Additionally, all models trained on FE-MNIST experienced drops in adversarial accuracy of 10 - 20\% when the evaluation is conducted with 40 random restarts and 100 PGD iterations. This is a larger drop then expected and is still an area being investigated as to the underlying factor. 

Finally, with FE-MNIST not all clients fare equally.
As shown in Figure~2, there is a tendency that clients which achieve better normal accuracy tend to exhibit better robustness.

\begin{minipage}{\textwidth}
  \begin{minipage}[b]{0.48\textwidth}
    \includegraphics[trim=5mm 0mm 5mm 0mm, width=\textwidth]{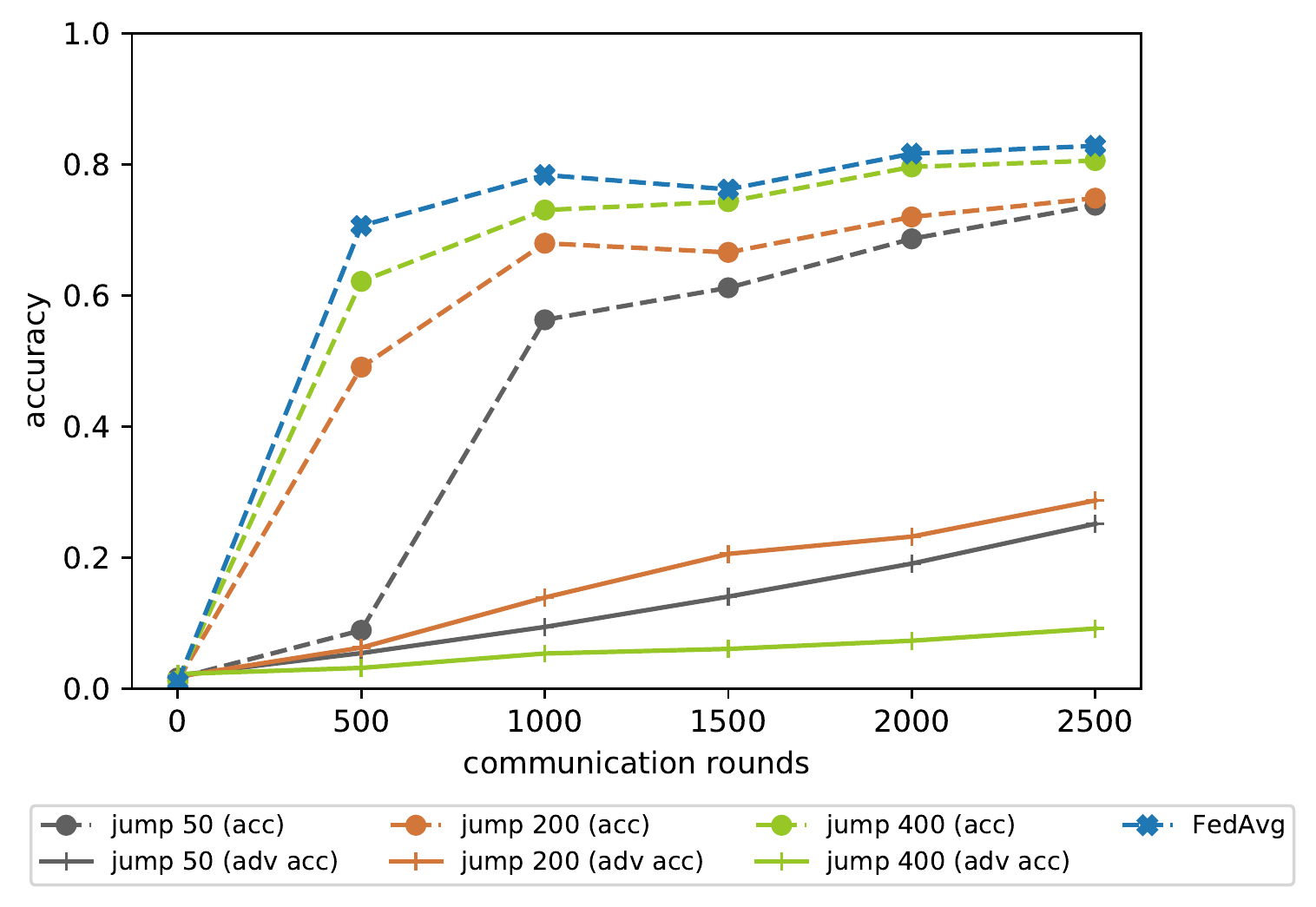}
    \captionof{figure}{Effect of changing $K/N$ at different times during the training of FE-MNIST.}
    \label{fig:jump}
    \end{minipage}
      \hfill
    \hfill
  \begin{minipage}[b]{0.48\textwidth}
    \centering
    \includegraphics[trim =0 0 0 0mm, clip,width=\textwidth]{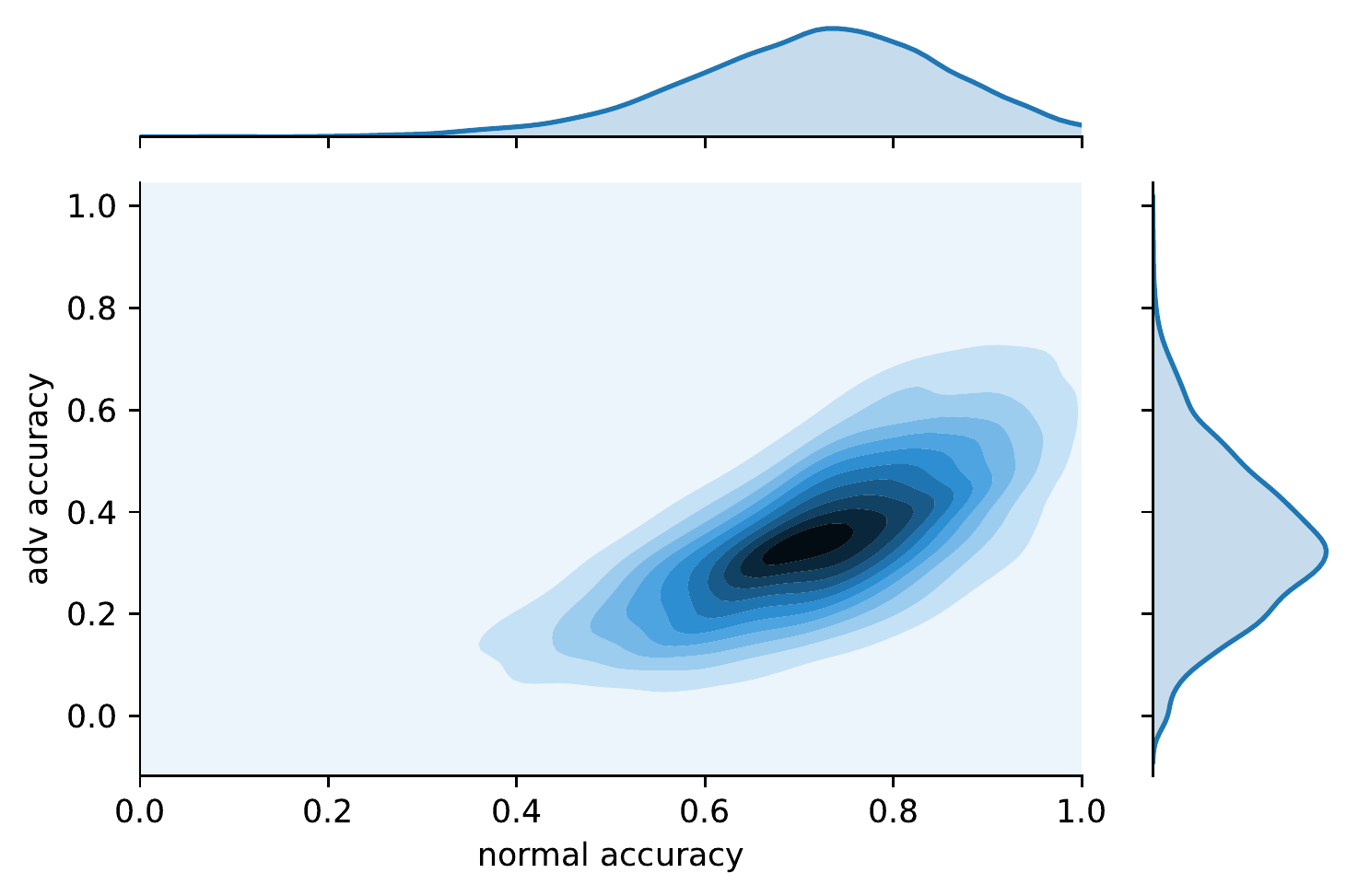}
    \label{fig:kde}
    \captionof{figure}{Distribution of individual client performance on local test data for FAT-0.8.}
  \end{minipage}
  \end{minipage}

\section{Byzantine Defences}

Secure aggregation of supplied updates is of significant importance in FL, else a single Byzanitne attacker can prevent global model convergence~\cite{blanchard2017machine}.
Therefore, adversarial training must be able to function alongside the pruning that secure aggregation schemes conduct.
We analyse three popular secure aggregation algorithms, namely Krum~\cite{blanchard2017machine}, Trimmed Mean, and Bulyan~\cite{mhamdi2018hidden}.
We examine each defense by measuring the performance in terms of regular and adversarial accuracy in the presence of malicious clients who are attempting to subvert the global model.
Due to computational constraints in performing FAT, we report results on MNIST and Fashion-MNIST.

\textbf{Krum}: Proposed in \cite{blanchard2017machine}, Krum selects a single update, $V$, that has the lowest score $s$ based on the squared distance to its $n-f-2$ neighbours where $n$ is the total number of clients in the system and $f$ is the number of permissible Byzantine clients,

\begin{equation}
    s(i) = \sum_{i \rightarrow j} || V_i - V_j ||^2.
\end{equation}

The sum is over the closest $n-f-2$ vectors compared to $i$. 
Krum uses the assumption that malicious updates are far from the benign ones in $L_2$ distance and hence are discarded during aggregation.

\textbf{Trimmed Mean}: The class of Trimmed Mean based defences \cite{xie2018generalized, yin2018byzantine, mhamdi2018hidden} work under the principle where for every dimension, $d$ of the parameter vector $V$ the median among the client updates is computed and only values close to the median are used in subsequent aggregation.
With $V_i^j$ referring to the $j^{\text{th}}$ dimension of update from client $i$, the  $j^{\text{th}}$ dimension of the final update, $W$, is computed as 

\begin{equation}
   W_j = \frac{1}{|U_j|}\sum_{i\in U_j} V_i^j.
\end{equation}

Here, $U_j$ defines the set of client updates whose updates are used in aggregation for the $j^{\text{th}}$ dimension.
In light of the comparison to previous works as noted in \cite{baruch2019alittle}, we implement the algorithm of \cite{mhamdi2018hidden} and define $U_j$ as the $(n-2f)$ closest updates to the median on dimension $j$.

\textbf{Bulyan}: The Bulyan defence \cite{mhamdi2018hidden} combines the two previously discussed defences.
Bulyan first iteratively applies Krum to produce a candidate set of $n - 2f$ updates.
Then this selection set is fed to a Trimmed Mean aggregation rule.
The advantage of this setup is that it uses Krum to filter updates that are far away from a $L_2$ perspective and it is proven to converge.
Moreover, Krum's shortcoming in its inability to capture small changes to many weights compared to large changes to a few weights is complemented by the  dimension wise pruning of the Trimmed Mean based class of defences.\footnote{A defence that we do not evaluate is DRACO \cite{chen2018draco} which is designed around \textit{distributed} rather than federated learning so data in DRACO is replicated across clients providing redundancy.
As privacy of client data is a requirement in FL DRACO cannot be used as widely as the defences we do evaluate.}

\subsection{Attacking Trimmed Mean and Bulyan}

We apply the convergence attack in \cite{baruch2019alittle} to the Trimmed Mean and Bulyan defences.
The intuition behind this attack relies on two observations.
First, is that many defenses assume that the attacker will supply updates which are far from the benign clients under a specific metric.
Second, is that the model updates supplied by clients will have a natural spread of values, which can be modelled as sampled from a normal distribution with mean $\mu$ with variance $\sigma$.
Building on these two observations the attacker tries to control which updates are selected by supplying updates of the form $\mu + k\sigma$, where $k$ determines how far from the mean to set the malicious updates.
The design of this attack strategy will favour the selection of malicious updates over those of benign clients updates which lie further away from the mean than $\mu + k\sigma$.
By consistently applying such updates an attacker can attempt to prevent convergence. In our experiments we set $k=-1.5$ for Trimmed Mean and $k=-1$ for Bulyan.

\subsection{Attacking Krum}

As Krum picks a single update we can pursue more ambitious attacker goals.
Here, the attacker introduces gradient masking into the global model.
This means that for an evaluation with white box PGD, the global model will appear robust and therefore a defender can be fooled into believing that the model is secure. 
However, gradient masking defences are generally brittle~\cite{athalye2018obfuscated}.
We introduce gradient masking via a model distillation attack where the attacker substitutes the robust PGD trained model that the defender is trying to achieve with one that is only robust due to defensive distillation~\cite{papernot2016distillation}.

\textbf{Distillation Attack:} The attacker conducts local defensive distillation by using the supplied global weights and applying a temperature $T=100$ to the softmax.
This model is trained on the local data and is then used to re-label the data producing soft labels.
Then, a new model initialised with the global weights and a similar temperature of $T=100$ on the softmax.
It is trained on the local data using the soft labels and the corresponding weights are sent as updates.
Empirically, adjusting all the weights using defensive distillation training scheme resulted in the supplied updates that differ too strongly and fail to surpass the barriers imposed by Krum.
Instead, we only train a single layer with the fewest weights.
Then, when attacking the defensively distilled model we report results when either using a black box transfer attack, or by using the principle in \cite{carlini2016defensive} where we scale the logits prior to taking the gradient to craft an adversarial example. 

This attack was ineffective against Trimmed Mean and Bulyan defences as due to the averaging around the median the fine grained control over the weights required in this attack is not maintained.

\section{Experiments}

\subsection{Experiment Setup}
We present results on MNIST and Fashion-MNIST datasets.
We replicate the setup in \cite{baruch2019alittle} where the FL system is composed of 51 clients and the updates are sent after training for one batch.  
Our setup differs in the use of Adam for local optimisation and the clients send the weight updates to the central server rather than the gradients. We also use a more standard batch size of 64 rather than 83 in \cite{baruch2019alittle}.
This is the same setup as we used for the MNIST and Fashion-MNIST results in Table \ref{tab:scale_results}. 
For each of these analyses we use the maximum amount of compromised clients the defence is designed to resist. For our setup of 51 clients this corresponds to 24 for Krum and Trimmed Mean and 12 for Bulyan.

\subsection{Results}

\begin{figure}
    \centering
    \includegraphics[trim=10mm 0 0mm 0, clip, scale=0.5]{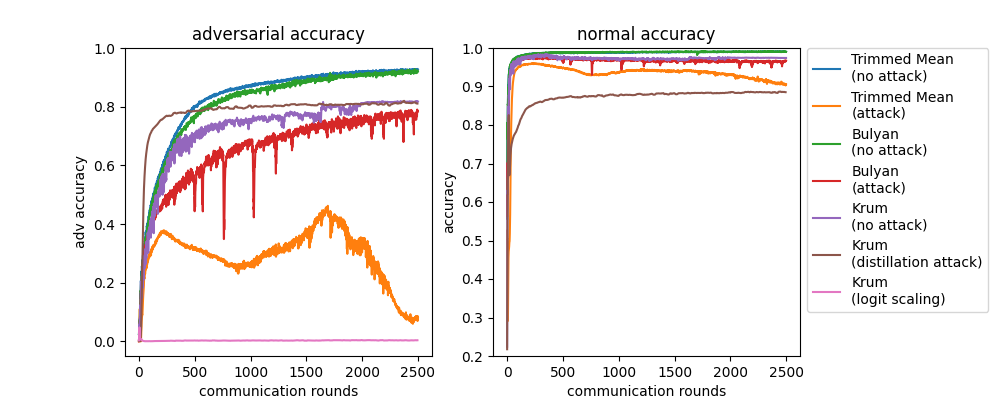}
    \caption{Training curves for various setups. We have the adversarial and normal performance for MNIST when conducting adversarial training with different Byzantine defences.}
    \label{fig:MNIST_Training_Curves}
\end{figure}
The results for our experiments in testing various defences is summarised in Table \ref{tab:byz_results}.
In the absence of attackers, the Trimmed Mean and Bulyan defences perform well with only modest drops in performance compared to standard federated averaging.
Krum, however, for the MNIST case experienced a drop in both normal and adversarial performance. 

When the models are subject to a Byzantine attacker the MNIST normal accuracy scores are broadly maintained under the convergence attack.
However, the Trimmed Mean defence is unable to maintain adversarial robustness when attacked, achieving a peak adversarial robustness of 46.20\%. 
But as we can see from Figure \ref{fig:MNIST_Training_Curves}, that level of performance is in itself tenuous as the robustness quickly drops.
The Bulyan defence offers the strongest level of adversarial accuracy under a Byzantine attacker with an accuracy of 60.09\% for Fashion-MNIST and 79.01\% for MNIST.

When we use our distillation attack on Krum, strong standard PGD performance is achieved on both Fashion-MNIST and MNIST datasets.
However, in both cases the actual model robustness to more specialised attacks which sidestepped the gradient masking is very low.
Using black box transfer attacks the mean robustness is 12.64\% and 33.74\% for MNIST and Fashion-MNIST respectively.
Even worse, if we use the logit scaling attack \cite{carlini2016defensive} the scores are 0.29\% and 4.19\% for MNIST and Fashion-MNIST.
The propensity of Krum for easy subversion with gradient masking gives a strong argument to discourage its use when seeking to achieve adversarial robustness. 

\begin{figure}
    \centering
    \includegraphics[trim=10mm 0 0mm 0, clip, scale=0.5]{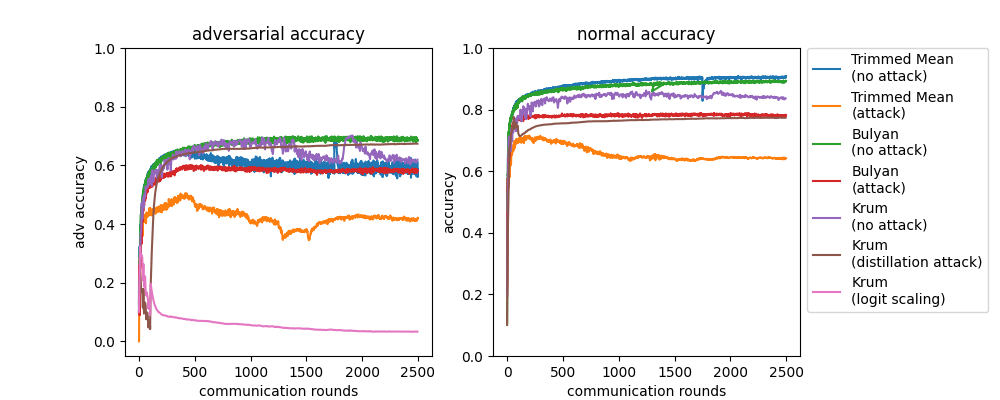}
    \caption{Training curves for various setups. We have the adversarial and normal performance for Fashion-MNIST when conducting adversarial training with different Byzantine defences.}
    \label{fig:FMNIST_Training_Curves}
\end{figure}

\begin{table}[h!]
    \centering
    \begin{tabular}{  c c  c  c c}
        \toprule
         Dataset  & Defence      & Attack & Adv. Acc. & Acc.\\
         \midrule
                & None         & None   &  90.92 & 99.09  \\
                \addlinespace[0.1cm]
                & \multirow{2}{*}{Trimmed Mean} & None   & 92.82 & 99.03 \\
                &                         & convergence   &  46.20 & 93.98 \\
                \addlinespace[0.1cm]
         MNIST  & \multirow{2}{*}{Buylan} & None   & 92.71 &  99.06 \\
                &                         & convergence   & 79.01 & 96.57 \\
                \addlinespace[0.1cm]
                & \multirow{2}{*}{Krum} & None &  81.82 & 97.41 \\
                &                       & \multirow{2}{*}{Distillation} & Apparent: 81.99 & \multirow{2}{*}{88.42} \\
                &                       &                               & Logit Scaled: 0.29    &  \\

        \midrule
                 & None         & None   &  69.41 & 88.67 \\
                 \addlinespace[0.1cm]
                 & \multirow{2}{*}{Trimmed Mean}  & None   &  69.07 & 85.07\\
         Fashion &                                & convergence & 50.54 &  70.02 \\
         MNIST   & \multirow{2}{*}{Buylan} & None & 69.97 & 88.94 \\
                 &                         & convergence & 60.09 & 78.09 \\ 
                 \addlinespace[0.1cm]
                 & \multirow{3}{*}{Krum} & None & 70.02 & 85.49 \\
                 &                       & \multirow{2}{*}{Distillation} & Apparent: 67.41 & \multirow{2}{*}{77.33} \\
                 &                       &                                & Logit Scaled: 4.19    &  \\
                                        
         \bottomrule
    \end{tabular}
    \caption{Results for PGD adversarial training performance when interacting with Byzantine defences, and when subject to different attacks. The results for no attack and defence are repeated here from Table \ref{tab:scale_results} for convenience. For the Krum distillation attack we give the models apparent PGD robustness when evaluating it with standard adversarial attacks and its robustness to a logit scaled attack \cite{carlini2016defensive}. With the logit scaled attack we report the mean performance after the distillation substitution attack has stabilised, 15 rounds in the case of MNIST and 100 rounds in the case of Fashion-MNIST. For all other results we report the scores when the maximum adversarial accuracy was achieved.}
    \label{tab:byz_results}
\end{table}

\section{Conclusion}

In this work we examined the effectiveness, scalability, and robustness of adversarial training in a federated learning system and demonstrated key open questions with federated adversarial training to guide future research efforts.
When using datasets constructed for federated learning \cite{caldas2018leaf} adversarial robustness comparable to the centralised setting can be achieved, however it shows high sensitivity to multiple hyperparameters.
It is important to note that any form of surgical hyperparameter search isn’t feasible in all federated learning setups as data is not readily accessible to begin with.
In the context of Byzantine attack strategies, only the Buylan defence offered meaningful defensive performance to maintain robustness.
Trimmed Mean suffered significant adversarial performance drops, and Krum could be subverted and mislead the defender entirely about the true robustness of the system.

This investigation has opened several open challenges for future work.
Firstly, there is a strong sensitivity to hyperparameters that we note across all experiments: from the ratio of adversarial examples in a batch to the amount of local computation clients conduct in each communication round.
This serves as a clear imperative for systematic evaluation the hyperparameter landscape and a need to develop more universally applicable training routines. 
Secondly, these experiments have further highlighted the gap between idealised and realistic FL scenarios \cite{caldas2018leaf}.
However, even then further considerations such as drift in data over time and issues with the potential lack of labelled local data or data mis-labeling \cite{jeong2020federated} can further complicate the path to adversarial robustness in FL systems.

\begin{ack}
This project has received funding from the European Union's Horizon 2020 research and innovation programme under grant agreement No 824988. https://musketeer.eu/. Additionally, this material is partially based upon work supported by the Defense Advanced Research Projects Agency (DARPA) under Contract No. HR001120C0013.
\end{ack}

\bibliography{refs.bib}
\bibliographystyle{plain}

\end{document}